\documentclass[Letter, 10 pt, conference]{ieeeconf}  % Comment this line out
\IEEEoverridecommandlockouts        % This command is only
\renewcommand{\baselinestretch}{1}

\title{\LARGE \bf Identification of stable models via nonparametric prediction error methods}

\author{Diego Romeres \and Gianluigi Pillonetto \and Alessandro Chiuso}

\usepackage{graphicx,color}
\usepackage{amsmath}
\usepackage{amssymb}
\usepackage{mathrsfs}
\usepackage{subfigure}
\usepackage{cite}
\usepackage{booktabs}

\newtheorem{problem}{Problem}
\newtheorem{remark}{Remark}

\newtheorem{alg}{Algorithm}

\newcommand{\norm}[1]{\left\lVert{#1}\right\rVert}                     		% \norm{x} -->  ||x|| with proper spacing
\newcommand{\kronecker}{\raisebox{1pt}{\ensuremath{\:\otimes\:}}} 
\newcommand{\E}{\mathbb{E}}
\newcommand{\C}{\mathbb{C}}
\newcommand{\R}{\mathbb{R}}
\newcommand{\Z}{\mathbb{Z}}
\newcommand\oprocendsymbol{\hbox{$\square$}}
\newcommand\oprocend{\relax\ifmmode\else\unskip\hfill\fi\oprocendsymbol}

\newcommand{\real}{\mathbb{R}}

\begin{document}
\maketitle
\thispagestyle{empty}
\pagestyle{empty}

%%%%%%%%%%%%%%%%%%%%%%%%%%%%%%%%%%%%%%%%%%%%%%%%%%%%%%%%%%%%%%%%%%%%%%%%%%%%%%%%
\begin{abstract}
A new Bayesian approach to linear system identification has been proposed in a series  of recent papers. The main idea is to frame linear system identification as predictor estimation in an infinite dimensional space, with the aid of regularization/Bayesian techniques.   This approach guarantees the identification of stable predictors based on the prediction error minimization. Unluckily, the stability of the predictors does not guarantee the stability of the impulse response of the system. In this paper  we propose and compare various techniques to address this issue. Simulations results comparing these techniques will be provided. 
%
%. First, we consider the so-called ``LMI - constraint'' approach and adapt it to  constrain the eigenvalues of the estimated model within the unit circle. Second, add to the ``classic''  %Stable-Spline algorithm a penalty term, depending on the maximum absolute value of the eigenvalue of the system. Third, we considered a Monte Carlo Markov Chain approach %sampling in both the space of hyper-parameters and of impulse responses. Simulations results comparing these techniques will be provided. 
\end{abstract}

%%%%%%%%%%%%%%%%%%%%%%%%%%%%%%%%%%%%%%%%%%%%%%%%%%%%%%%%%%%%%%%%%%%%%%%%%%%%%%%%

% -------------------------------------------------------------------------------------- 
\section{Introduction}

Recent approaches for linear system identification describe the unknown system directly in terms of impulse response, thus describing an infinite dimensional model class. 
 Needless to say, this is not entirely free of difficulties, since an alternative way to control the model complexity, i.e., to face the so called-bias variance tradeoff  \cite{Ljung,Soderstrom}, need to be found. It has been shown in the recent  literature that the apparatus of Reproducing Kernel Hilbert Spaces (RHKS) or, equivalently, Bayesian Statistics  provide powerful tools to face this tradeoff. 

The paper \cite{GP-AC-GdN:11} has shown how these infinite dimensional model classes can be used for  identification of linear systems in the framework of prediction error methods, leading naturally to stable predictors. Yet stability of the predictor model does not necessarily guarantee stability of the so called ``forward'' (or simulation) model. As a matter of fact, we faced this stability  issue when performing identification on a real data set from EEG recordings. Physical insight in this case suggests that the transfer function describing the link between potentials in different brain locations are expected to be stable, while the identified models where not. 

Therefore, motivated by this real-world application, in this paper we shall tackle the problem of identifying stable (simulation) models when nonparametric prediction error methods \cite{GP-AC-GdN:11} are used. We shall describe and compare, through an extensive simulation study, four possible solutions to this problem. 

The paper is structured as follows: Section \ref{Section: Statement of the problem} formulates the problem.  Sections \ref{Section: Stabilization with LMI}-\ref{Section: MCMC } introduce  four different approaches to guarantee  stability of the   identified  models. Experimental  results are described in Section \ref{Section: Simulations} and  conclusions are drawn in Section \ref{Section: Conclusions}.
%\vspace*{-1.5 mm}
%
\subsubsection*{Notation} Given a matrix $M$, $M^\top$ shall denote its transpose, $\sigma(M) $ will be its eigenvalues. If $A(z)$ is a polynomial, $\sigma(A(z))$ will denote the set of  roots of $A(z)$. Given two discrete time  jointly stationary stochastic process $y(t)$ and $z(t)$, the symbol $\E[y(t) | z(s), s<t]$ shall denote the linear minimum variance estimator (conditional expectation in the Gaussian case) of $y(t)$ given the past $(s<t)$ history of $z(t)$. 

 %-------------------------------------------------------------------------------------- %
\section{Statement of the model stabilization problem}
\label{Section: Statement of the problem}
We shall consider two   jointly stationary discrete time zero mean stochastic processes $\{u(t)\},\{y(t)\}$, $t\in \Z$, respectively the   ``input''  and ``output'' processes.
% Actually this nomenclature is completely arbitrary since we shall make no assumptions on the possible presence of feedback from $\{u(t)\}$ to $\{y(t)\}$, so that it is perfectly legitimate to consider a ``feedback'' path where $\{y(t)\}$ plays the role of the input and $\{u(t)\}$ that of the output.

As shown in \cite{SoedestroemLG1976,NgGA1977} under these assumptions there is an essentially unique representation of $y(t)$ in terms of $u(t)$ of the form 
\begin{equation}\label{IOmodel}\begin{array}{rcl}
y(t) &= &P(z) u(t) + H(z) e(t)\\
e(t)&:=& y(t) - \E[y(t) | y(s),u(s), s<t]% \\
%& =& y(t) - \hat y(t|t-1)
\end{array}\end{equation}
\vspace*{-2mm}
where
\vspace*{-1mm}
\begin{equation}\label{TF}
P(z):=\sum_{k=1}^{\infty} p_k z^{-k} \quad \quad H(z):=\sum_{k=0}^{\infty} h_k z^{-k}  \quad h_0=1
\end{equation}
and $H(z)$ is minimum-phase. This  guarantees stability of the predictor $\hat y(t|t-1):=\E[y(t)|y(s),s<t]$:
\begin{equation}\label{predictor}
\hat y(t|t-1) = H(z)^{-1}\left[ (H(z) -1)y(t) + P(z) u(t)\right]
\end{equation}
%\textbf{To simplify the notation we define $p:=\{p_k\}_{k\in \Z^+}$ and $h:=\{h_k\}_{k\in \Z^+}$ which will be used in the remaining of the paper}
In this paper we shall also assume that $P(z)$ (and thus $H(z)$) are stable\footnote{Note that in a feedback configuration $P(z)$ is in principle allowed to be unstable provided there is a stabilizing feedback in action.} (i.e., analytic inside the open unit disc).

Prediction error approaches to system identification \cite{Ljung,Soderstrom} are based on estimating  the predictor model \begin{equation}\label{predictor}
\begin{array}{c}
\hat{y}(t\vert t-1) = F(z)y(t)+ G(z)u(t) \\
F(z) = \sum_{k=1}^{\infty} f_k z^{-k} \quad G(z) = \sum_{k=1}^{\infty} g_k z^{-k} 
\end{array}
\end{equation}
Classic parametric methods \cite{Ljung,Soderstrom}  start from a parametric description $P_\theta(z)$ and $H_\theta(z)$ of $P(z)$ and $H(z)$ in \eqref{IOmodel}. This parametrization is usually constrained ($\theta \in \Theta$) so as to account for prior knowledge such as stability of $P_\theta(z)$, $H_\theta(z)$ and $H_\theta^{-1}(z)$. This induces a natural parametrization of the predictor $\hat y(t|t-1)$ which is  thus denoted by $\hat y_\theta(t|t-1)$. Given a data set $y:=\{y(t)\}_{t=1,..,T}$, $u:=\{u(t)\}_{t=1,..,T}$, the parameters $\theta$ are then estimated minimizing the squared loss
\begin{equation}\label{PE}
\sum_{t=1}^T \left(y(t) - \hat y_\theta(t|t-1)\right)^2.
\end{equation}
More recently prediction error identification has been formulated in a nonparametric framework \cite{GP-AC-GdN:11}. The main issue working in a nonparametric (possibly infinite dimensional) framework is that the problem of finding estimators $\hat f,\hat g$ of $f:=\{f_k\}_{k\in \Z^+}$ and $g:=\{g_k\}_{k\in \Z^+}$ from measurements $y,u$ is an ill-posed inverse problem \cite{Tikhonov}. The main idea, borrowed from \cite{SS2010} is to minimize the prediction error \eqref{PE} searching for $\{f_k\}_{k\in \Z^+}$ and  $\{g_k\}_{k\in \Z^+}$ in a suitable Reproducing Kernel Hilbert Space (RKHS) \cite{Aronszajn50} which acts as a regularizer, also encoding   notions of ``stability'' of the predictor (e.g. making sure that the estimated $\hat F(z)$ and $\hat G(z)$ are BIBO stable with probability one), see \cite{GP-AC-GdN:11,SS2010} for details. 
Equivalently one can think that $\{f_k\}_{k\in \Z^+}$ and  $\{g_k\}_{k\in \Z^+}$ are modeled as independent zero mean Gaussian Process \cite{Rasmussen} with a suitable covariance  $K(t,s) = cov(f_t,f_s) = cov(g_t,g_s)$ (the same as the Reproducing Kernel above). This covariance is usually parametrized by some unknown hyperparameters $\eta$, which will be made explicit in the notation using a subscript, e.g.   $K_\eta $ and  $p_\eta(f,g)=p_\eta(f)p_\eta(g)$. 
Under the assumption that the innovation process is Gaussian and independent of $f=\{f_k\}_{k\in \Z^+}$ and  $g=\{g_k\}_{k\in \Z^+}$,  also  the marginal $p_\eta(y,u)$
%\begin{equation}\label{ML}
%p_\eta(y,u): =\int p(y,u|f,g)p_\eta(f,g)\,dg\,df
%\end{equation}
and the  posterior $p_\eta(f,g|y,u)$
%\begin{equation}\label{posterior}
%p_\eta(f,g|y,u):= \frac{p(y,u|f,g)p_\eta(f,g)}{p_\eta(y,u)}
%\end{equation}
are Gaussian, see \cite{GP-AC-GdN:11} for details. The marginal density $p_\eta(y,u)$, also called marginal likelihood, can be used to estimate the unknown hyperparameter as:
\begin{equation}\label{maxML}
\hat \eta_{ML}:= {\rm arg\; max}_\eta \, p_\eta(y,u).
\end{equation}
Then, following the Empirical Bayes paradigm, 
  estimators of $f=\{f_k\}_{k\in \Z^+}$ and  $g=\{g_k\}_{k\in \Z^+}$ are then found from their posterior density $p_\eta(f,g|y,u)$ having fixed the hyperparameters to their estimated value $\hat\eta_{ML}$  \cite{GP-AC-GdN:11}:
  \begin{equation}\label{EB}
  \hat f:= \E_{\hat\eta_{ML}}[f|y,u], \quad  \hat g:= \E_{\hat\eta_{ML}}[g|y,u]
  \end{equation}
  where $\E_{\hat\eta_{ML}}[\cdot|\cdot]$ denotes conditional expection having fixed  $\eta=\hat\eta_{ML}$.
%It is a well known fact in literature, see e.g. \cite{GP-AC-GdN:11}, that it is possible to estimate a BIBO stable one-step-ahead predictor of $y(t)$ by imposing properties to the a priori covariance of the impulse responses. Accordingly, let the predictor takes the form:
%\begin{equation}
%\hat{y}(t\vert t-1) = \displaystyle \sum_{k=1}^\infty f_kz^{-k}y(t)+ \sum_{k=1}^\infty g_k z^{-k}u(t) 
%\end{equation}
%where the impulse responses $f(t),g(t)$ are guaranteed to be BIBO stable. 

Unfortunately,  BIBO stability of the impulse responses of $\{\hat f_k\}_{k\in \Z^+}$ and  $\{\hat g_k\}_{k\in \Z^+}$ does not guarantee  BIBO stability of  the estimates
\begin{equation}
\label{eq: system estimated from the predictor}
%\hat P(z):= \frac{ \displaystyle \sum_{k=1}^\infty \hat g_k(\eta)z^{-k}}{1- \displaystyle \sum_{k=1}^\infty \hat f_k(\eta)z^{-k}}
%\quad \hat H(z) := \frac{1}{1-\displaystyle \sum_{k=1}^\infty \hat f_k(\eta)z^{-k}}
\hat P(z):= \frac{ \hat G(z)}{1- \hat F(z)  },
\quad \hat H(z) := \frac{1}{1- \hat F(z)}
%\hat P(z):= \frac{ \displaystyle \sum_{k=1}^\infty \hat g_kz^{-k}}{1- \displaystyle \sum_{k=1}^\infty \hat f_kz^{-k}}
%\quad \hat H(z) := \frac{1}{1-\displaystyle \sum_{k=1}^\infty \hat f_k z^{-k}}
\end{equation}
of $P(z)$ and $H(z)$ in \eqref{IOmodel}. In fact, BIBO stability of the sequences $\{\hat f_k\}_{k\in \Z^+}$ and  $\{\hat g_k\}_{k\in \Z^+}$ have no  relation with stability of $\hat P(z)$ and $\hat H(z)$ which, if no cancellations occur, depends on the  zeros  of $1-  \hat F(z)=1- \sum_{k=1}^\infty \hat f_kz^{-k}$.\\
%\drmargin{Possiamo dire che noi per semplicità consideriamo sempre che non si siano cancellazioni, cosi mi viene piu semplice usare i concetti di autovalori, poli e modi con lo stesso significato}
%
For practical purposes when estimating the predictor model \eqref{predictor}, the impulse responses $\{f_k\}_{k\in \Z^+}$ and  $\{g_k\}_{k\in \Z^+}$ are truncated to a finite (yet arbitrarily large) $p$, so that we~assume
$\label{FIR}
F(z) = \sum_{k=1}^{p} f_k z^{-k}, \, G(z) = \sum_{k=1}^{p} g_k z^{-k} 
$.

Thus, the problem we consider in this paper, can be formulated as follows:

\begin{problem}
\label{prb: original} 
Given  $y(t),u(t)$, $t\in [1,T]$, find  $\{\hat f_k\}_{k\in [1,p]}$ and  $\{\hat g_k\}_{k\in [1,p]}$ so that 
$\hat P(z)$ and $\hat H(z)$  in \eqref{eq: system estimated from the predictor} are BIBO stable transfer functions. A sufficient generic\footnote{i.e., if no cancellations occur, which is generic for estimated impulse responses.} condition for this to happen is that
\begin{equation}
\begin{array}{c}
\label{eq: A(z) model denominator}
A(z) = z^p(1-\sum_{k=1}^p \hat f_kz^{-k})=z^p -  [z^{p-1} \ldots 1] \hat {f} \\
\hat f:=[\hat f_1, \hat f_2,\dots,\hat f_p]^\top
\end{array}
\end{equation}
is stable, i.e., has all roots inside ${\cal D}:=\{z \in \C: |z|<1 \}$.
%the open unit disc of the complex plane $\C$.
\end{problem}

In the following we describe and compare three different techniques to achieve this aim. For each technique Problem~\ref{prb: original} is properly reformulated.
{In order to simplify the notation, in what follows, the input $u$ will be dropped from the notation; therefore, for instance, we shall use $p_\eta(y)$ in lieu of $p_\eta(y,u)$.}
%
%\begin{problem}
%\end{problem}

% -------------------------------------------------------------------------------------- %

% -------------------------------------------------------------------------------------- %
\section{Stabilization via LMI constraint}
\label{Section: Stabilization with LMI}
% -------------------------------------------------------------------------------------- %
The first stabilization technique is based on formulating stability of the  model \eqref{eq: system estimated from the predictor}  as a constraint on the eigenvalues of the companion matrix of $A(z)$ in \eqref{eq: A(z) model denominator}. This constraint can be characterized in terms of  Linear Matrix Inequalities (LMI) as discussed in \cite{CM-GP:96}, and used later on in \cite{DNM-RAC:13} to enforce stable models in subspace identification, thus leading to: 
\begin{problem}[Reformulation] Given a preliminary estimate $\tilde f:=[\tilde f_1,..,\tilde f_p]^\top$, find a vector of coefficients $\hat f$ so that
\label{prb: LMI}
\begin{equation}\label{CONV}
\hat{f}=\arg \min_{f \in {\cal F}_{\cal D}} \norm{f- \tilde{f}}^2
\end{equation}
where ${\cal F}_{\cal D}:=\{ f \in \R^p : |\lambda|<1 \; \forall \lambda \; s.t.\;  A(\lambda)=0, A(z)= z^p -  [z^{p-1} \ldots 1] \hat{f}\}$, can be  described by an LMI constraint as discussed below.
\end{problem}
%Problem \eqref{CONV} is a quadratic program with LMI constraints which can be easily solved by available software \textit{CVX Toolbox} in Matlab \cite{MG-SB-YY:06}.\\
%The estimators $\tilde f$ which, in principle, can be any estimator of $f$, is obtained from the Bayesian approach described in \cite{GP-AC-GdN:11}%\cite[Section 4]{DNM-RAC:13}
%\drmargin{ho cambiato reference}
%, which will be denoted as $\tilde f:=\tilde f_B$.

%One might question why should we measure how well $\hat f$ fits $\tilde f$ in terms of 2-norm. 
It should be observed that the use of the 2-norm in  \eqref{CONV} is  entirely arbitrary and, in fact, considering some form of model approximation error (e.g. difference of  output predictors) would be preferable. In addition, when  $\tilde f$ is the outcome of a preliminary estimation step, a  principled solution would require  accounting for the distribution of $\tilde f$.

 However, this brings in some technical difficulties related to the formulation of the quadratic problem, therefore, it is still subject of research. 

\subsection*{Formulation of the LMI constraint}
%\begin{equation}
%{\cal D} = \{z \in \complex : \vert z \vert < 1\}
%\end{equation}
%It is well known \cite{CM-GP:96} that the open unit disc ${\cal D}$ can be expressed in terms of the matrix polynomial
%\vspace*{-2mm}
%\begin{equation}
%f_{\cal D}(z)= I_2+
%\begin{bmatrix}
%0 & 1 \\
%0& 0
%\end{bmatrix} z
%+
%\begin{bmatrix}
%0 & 0 \\
%1& 0
%\end{bmatrix} \bar{z}
%\end{equation}
%%
%as ${\cal D} = \{ z \in \C \; s.t. \; f_{\cal D}(z) > 0 \} $,\\

As shown in \cite{CM-GP:96}, a matrix $F$ has all its eigenvalues   in the LMI region ${\cal D} = \{ z \in \C \; s.t. \; f_{\cal D}(z) > 0 \} $, where $f_{\cal D}(z)$ is an opportune polynomial matrix, if and only if there exists $P=P^\top \geq 0 $ s.t. 
\begin{equation}\label{LMI_constraint}
M(F,P)=I_2 \otimes P + 
\left(
\begin{bmatrix}
0 & 1 \\
0& 0
\end{bmatrix} \otimes (FP) 
\right)
+
\bigg( * \bigg)^\top \geq 0 
\end{equation}
%\begin{equation}\label{LMI_constraint}
%M(F,P)=I_2 \otimes P + \begin{bmatrix}
%0 & 1 \\
%0& 0
%\end{bmatrix} \otimes (FP) 
%+
%\begin{bmatrix}
%0 & 0 \\
%1& 0
%\end{bmatrix} \otimes (FP)^\top \geq 0 
%\end{equation}

According to \cite[Theorem 1]{DNM-RAC:13}, which presents small variations w.r.t the original central theorem in \cite{CM-GP:96}, we define the  companion matrix of ${f}$ as $\Psi(f) \in \real^{p \times p}$. Therefore, using  \eqref{LMI_constraint}, $f$ is (Schur) stable if and only if $\exists P=P^\top \geq0$ such that $M(\Psi(f) ,P) \geq 0$.
%\drmargin{si puo anche togliere}
%$$\Psi(f) := \left[\begin{array}{c|ccc}
%0 &  &  &  \\
%\vdots &  & I_{p-1} &  \\
%0 &  &  & \\\hline
%-f_p & -f_{p-1} & \dots & -f_1
%\end{array}\right] =
%\begin{bmatrix}
% J\\\midrule
% f
%\end{bmatrix} \in \real^{p \times p}$$
%$$\Psi(f) := \left[\begin{array}{ccccc}
%
%0 & 1 & 0 &  \dots & 0 \\
%0 & 0 & 1 & \dots & 0 \\
%\vdots & \vdots & \vdots & \ddots  & \vdots\\
%0 & 0 & 0 & \dots & 1\\
%-f_p & -f_{p-1} & -f_{p-2} & \dots & -f_1
%\end{array}\right] \in \real^{p \times p}$$

%The LMI constraint to guarantee the stability of $\Psi(f) $ and therefore of $f$ is $M(\Psi(f) ,P) \geq 0$, where $M(F,P)$ is defined in \eqref{LMI_constraint}. 
Unfortunately $M(\Psi(f) ,P)$ this is not linear in $f$ and $P$ since their product appears. Similarly to \cite{DNM-RAC:13}, this calls for a reparametrization of the constraint as follows:    define   the vector 
%matrix $P \in \real^{p \times p}, \, \Psi \in \real^{p} $ such that 
$ \psi:= P {f} $ (so that $f = P^{-1} \psi$),  $J:= [\underline{0} \, I_{p-1}]$,  and  
$M(\psi,P):= M(\Psi(f),P)$
i.e., 
\begin{equation}
\label{eq: LMI constraint stability}
M(\psi,P) = I_2\kronecker P +
\left(
\begin{bmatrix}
0 & 1\\
0 &0
\end{bmatrix}
\kronecker
\begin{bmatrix}
JP\\
\Psi^T
\end{bmatrix}
\right)
+
\bigg(
*
\bigg)^\top
\end{equation}
which is  linear in $\psi$ and $P$. 
%Now the stability constraint  \eqref{LMI_constraint} can be rewritten as 
%Now, we are able to formulate the stability constraint using the LMI  constraint $M(\psi,P) \geq 0 $, where $M(\psi,P) $ has been defined in \eqref{eq: LMI constraint stability}. 
Thus problem~\ref{prb: LMI} can be reformulated as: 
\begin{align}
\label{eq: LMI optimization problem first step}
 \notag
\hat{\psi},\hat{P}= & \arg\min_{f,P} \norm{\psi-P \hat{f}_{B}} ^2\\
\text{s. t.} \quad &  M(\psi,P) \geq 0,  \quad Tr(P) =  p, \quad   P = P^T \geq 0 &%\\% \notag
%&  Tr(P) = p\\ \notag
%& \qquad P = P^T \geq 0
\end{align}
where the constraint $Tr(P)=p$ is added to  improve  the numerical conditioning, see \cite{DNM-RAC:13} for further details.\\
The solution $\hat f$ of   Problem~\ref{prb: LMI} is finally computed as:
\begin{equation}
\label{eq: LMI stable predictor first step}
\hat{f} = \hat{P}^{-1}\hat{\psi}
\end{equation}

{In the remaining of the paper the model $\hat P(z)$ obtained by plugging  in \eqref{eq: system estimated from the predictor}   the estimators $\hat f$  and $\hat g$ obtained  respectively from   \eqref{eq: LMI stable predictor first step} and the Bayesian procedure in \cite{GP-AC-GdN:11}, will be called ``LMI'' model.}
\section{Stabilization via Penalty Function}
\label{Section: Stabilization with Smooth Barrier}
% -------------------------------------------------------------------------------------- %

The second stabilization technique is formulated to act directly inside the Bayesian procedure. As briefly discussed in section~\ref{Section: Statement of the problem}, a crucial step of the Bayesian procedure is the estimation of the hyperparameter vector $\eta$ through marginal likelihood optimization \eqref{maxML}. It is in principle possible to 
 restrict the set of admissible hyperparameters to a subset $\Xi_S$ which lead to estimators \eqref{EB}  corresponding to stable models $\hat P(z)$ and $\hat H(z)$.
%It turns out that some hyperparameters $\eta$ may lead to estimators \eqref{EB} which do not correspond to stable models $\hat P(z)$ and $\hat H(z)$. Thus, one possible remedy is to restrict the set of admissible hyperparameters to a subset $\Xi_S$ which lead to stable models. 
This is not entirely trivial as the estimators (and thus the set $\Xi_S$) depend on the measured data $y,u$.  
This leads to the following:
\begin{problem}[Reformulation]
\label{prb: penalty function}
Estimate the hyperparameters $\eta$ solving \begin{equation}
\label{eq: max mlik problem}
\hat{\eta} = \arg\max_{{\eta} \in \Xi_S}  p_\eta(y) =  \arg\min_{{\eta} \in \Xi_S} -\ln  p_\eta(y)
\end{equation}
to the set  $\Xi_S = \{ \eta \vert A(z) \text{ Stable } \}$, i.e., the set of hyperparameters which lead to stable models $\hat P(z)$, $\hat H(z)$. 
\end{problem}
To force $\eta \in \Xi_S$, we can add a penalty function to the criterion in  \eqref{eq: max mlik problem} which acts as a barrier  to keep the estimate $\hat\eta$ away  from the set of hyperparameters $\eta$  leading  to an unstable $A(z)$. In order to do so, we define $A_\eta(z)$ the polynomial $A(z)$ in  \eqref{eq: A(z) model denominator} built with the estimator 
\begin{equation}
\label{eq: f estimator depend eta}
\hat f_\eta:=\E_{\eta}[f|y,u],
\end{equation} 
and $\bar{\rho}_\eta = \max \vert \sigma(A_\eta(z))\vert$. Next define the  penalty~function: 
\vspace*{-2mm}
\begin{equation}
\label{eq: penalty function}
J(\bar{\rho}_\eta) = \frac{1}{(\alpha(\delta-\bar{\rho}_\eta))^\alpha} -\frac{1}{(\alpha\delta)^\alpha}
\end{equation}
where $\delta \geq 1$ is a scalar which defines the barrier, $\alpha$ is a positive scalar which adjust how steep the barrier is.

\begin{figure}[!h]
\begin{center}
\includegraphics[width=.8\columnwidth]{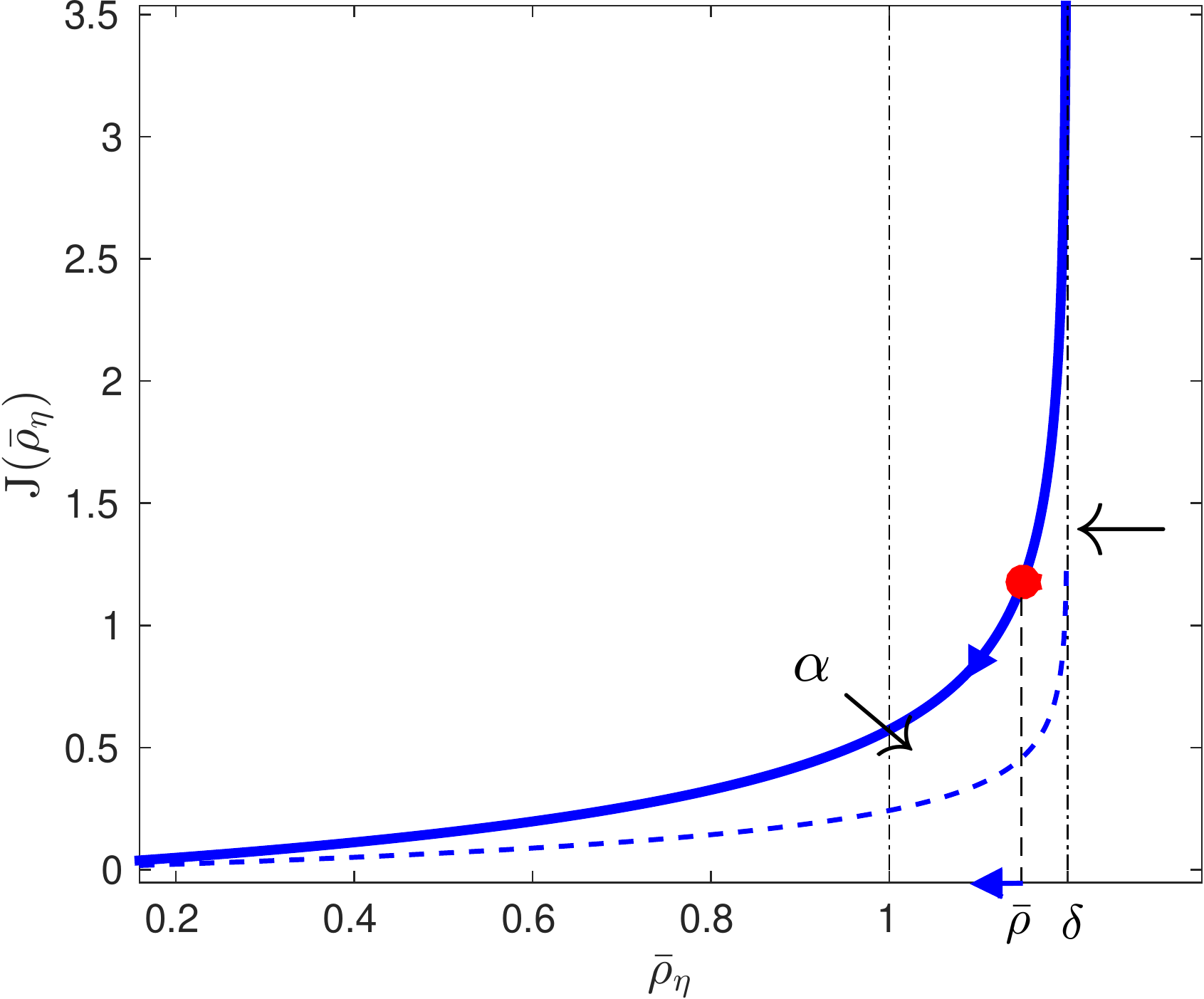}
\caption{Representation of the penalty function $J(\bar{\rho}_\eta)$. The red bullet represents the penalty function value associated to a specific $\bar{\rho}$ in an illustrative example. The blue (head filled) arrows show the effect of the penalty function on $\bar{\rho}$, the black (head no-filled) arrows the effects of changing the parameters $\alpha$ and $\delta$. The blue dashed line represents the variation of $J(\bar{\rho}_\eta)$ after a reduction of $\alpha$.}
\label{fig: penalty function}
\end{center}
\end{figure}
%
%\begin{itemize}
%\item $\bar{\rho}_{\eta} = \bar{\lambda}(A(z)) $
%\item $\delta =$ Imposed barrier
%\item $\lim_{\bar{\rho}_{\eta}\rightarrow \delta} J_p(\bar{\rho}_{\eta}) = \infty$, $\lim_{\bar{\rho}_{\eta} \rightarrow 0} J_p(\bar{\rho}_{\eta}) = 0$
%\item $\alpha$ adjust the steepness
%\end{itemize}
%
As we can see in  Figure \ref{fig: penalty function} function \eqref{eq: penalty function} diverges ($J(\bar{\rho}_\eta) \rightarrow \infty$) when $\bar{\rho}\rightarrow \delta$ and $J(\bar{\rho}_\eta) \rightarrow 0$ when $\bar{\rho} \rightarrow 0$.  Thus  when \eqref{eq: penalty function} is added to  the minimization problem \eqref{eq: max mlik problem}, the solution  $\bar{\rho}$ is pushed inside the stability region. % thanks to the offset term which is the value of \eqref{eq: cost function} evaluated in $0$. 
The parameters $\alpha$ and $\delta$ are iteratively adjusted so as to guarantee that the final solution leads to a stable model, i.e. solves the constrained problem \eqref{eq: max mlik problem}.

 Notice that when $\alpha \rightarrow 0$, $J(\bar{\rho}_\eta)$ gives no penalty for $\bar{\rho}_\eta<\delta$ and infinite penalty for $\bar{\rho}_\eta \geq \delta$. %tends the indicator function  to be a rotate step function with infinite value in $\delta$ and zero value in $[0,\delta[$.\\
Elaborating upon the intuition above, it is easy to prove that the solution of  Problem~\ref{prb: penalty function} can be found by the algorithm described below: 
%%%%%%%%%%%%%%%%%%%%%%%%%%%%%%%%%%%%%%%%%%%%%%%%%%%%%%%%%
\begin{alg}
\label{alg: penalty function}
\mbox{ }
\begin{enumerate}
\item Initialization:
\begin{itemize}
\item Compute $\eta_0$ using \eqref{eq: max mlik problem} and set $\alpha =1$.
\item Compute the predictor impulse response $\hat f_{\eta_0}$ using \eqref{eq: f estimator depend eta}, then determine the associated $A_{\eta_0}(z)$, $\bar \rho_{\eta_0}$.
%$ \eta_0 \rightarrow f_{\eta_0} \rightarrow A_{\eta_0} \rightarrow \bar{\rho}_{\eta_0}$
\end{itemize}
\item While $\bar{\rho}_{\eta_k} \geq 1$
\begin{itemize}
\item Set $\delta = \bar{\rho}_{\eta_k}(1+\epsilon)$ %($\delta$ cannot increase from this value)
\item Compute
\begin{equation}
\label{eq: min lml + penalty function}
\eta_k= \arg \min_{\eta} -\ln p_\eta(y) + J(\bar \rho_\eta)
\end{equation} 
%$\eta_k = \arg\min-p(y\vert \eta) + J_p(\eta)$
and the associated $\bar{\rho}_{\eta_k}$
\item If the value of $- \ln p_{\eta_k}(y) + J(\bar \rho_{\eta_k})$ is unchanged w.r.t. the $k-1$ iteration, then perform the update:\\
$\alpha =\alpha-\Delta \alpha, \delta =\delta-\Delta \delta$  where $\Delta \alpha$ and $\Delta \delta$ are chosen sufficiently small
\end{itemize}
\item Set $\alpha = \epsilon $ and $\delta=1$. \\
Finally, the solution of Problem~\ref{prb: penalty function} is given by:
\begin{eqnarray}
\hat{\eta} &=& \arg \min_{\eta}- \ln p_\eta(y) + J_p(\eta)\\
\label{eq: ML+PF stable predictor}
\hat{f}& =& \mathbb \E_{\hat\eta}[f\vert y,u], \quad \hat{g} = \mathbb \E_{\hat\eta}[g\vert y,u]
\end{eqnarray}
{In the remaining of the paper the model obtained by \eqref{eq: system estimated from the predictor} using \eqref{eq: ML+PF stable predictor} will be called ``ML + PF'' model.}
\end{enumerate}
\end{alg}
\begin{remark}
Notice that the iterative procedure which updates $\delta$ and $\alpha$ is needed because, in general, it is not guaranteed that one can find an initial value of $\eta \in \Xi_S$. 
Note also that the set $\Xi_S$ is always non-empty provided the hyperparameter vector $\eta$ includes a scaling factor for the Kernel,  i.e., a non negative scalar which multiplies the Kernel matrix. In fact, if this is the case, there exist values of $\eta$ which lead to  $\hat f=0$ which, in turn leads to stable $\hat P(z)$ and $\hat H(z)$. 

\end{remark}
%descrivere smooth barriera, descrivere a parole l algoritmo\\
%fminsearch\\
%avvicino barriera\\
%\begin{equation}
%\label{eq: derivative of cost function}
%f = \alpha^2(\alpha(\delta-\rho_{\max}))^{-\alpha-1}
%\end{equation}
%%%%%%%%%%%%%%%%%%%%%%%%%%%%%%%%%%%%%%%%%%%%%%%%%%%%%%%%%%%%%%%%%%%%%%

\section{Stabilization via Markov Chain Monte Carlo}
\label{Section: MCMC }
%%%%%%%%%%%%%%%%%%%%%%%%%%%%%%%%%%%%%%%%%%%%%%%%%%%%%%%%%%%%%%%%%%%%%%
In this Section we shall present a MCMC approach which yields  the so called \emph{full Bayes} estimator of $f$ and $g$, introducing a (possibly non-informative)  prior density \footnote{This may be a uniform distribution if the domain is compact.} $p(\eta)$  on the hyperparameter vector $\eta$. 
%To the purpose of marginalizing over $\eta$   we consider a non-informative prior.  
In order to enforce the stability constraint   we consider the ``stable'' posterior distribution
\begin{equation}
\label{eq: stable posterior general form}
\begin{array}{rcl}
p_S(f,g\vert y) &=& \frac{1}{p(y)}\int p(y|f,g)p_S(f,g|\eta) p(\eta)\, d\eta 
\end{array}
\end{equation}
where $p_S(f,g|\eta)$ is the ``truncated'' Gaussian prior
\begin{equation}\label{truncatedGaussian}
p_S(f,g|\eta): = \left\{ \begin{array}{cl} k_\eta p_\eta(f,g)& f:  A(z)\;\; {\rm stable} \\
0 & {\rm otherwise}\end{array}\right.
\end{equation}
which, a priori, excludes all impulse responses $f$ which lead to unstable $A(z)$. Note that the constant $k_\eta$ in \eqref{truncatedGaussian} equals
$
k_{\eta}: = \frac{1}{\int_{f\in {\cal F}} p(f,g|\eta)\, dfdg},$ where ${\cal F}:=\{f| A(z) \; {\rm stable}\}
$.
Unfortunately,  the ``stable'' conditional 
$$p_{S}(f,g|y,\eta): = \frac{p(y|f,g) p_S(f,g|\eta)}{p_S(y,\eta)}$$ is not Gaussian and, in addition, the integral in \eqref{eq: stable posterior general form} cannot be computed in closed form. Therefore we tackle the problem using MCMC methods:

\vspace{.1cm}
\begin{problem}[Reformulation]
\label{prb: MCMC}\mbox{}\newline
Obtain a sampling approximation of the ``stable'' posterior distribution \eqref{eq: stable posterior general form}. Compute from these samples the estimates $\hat f,\hat g$ in \eqref{EB} and $\hat P,\hat H$ in \eqref{eq: system estimated from the predictor} which satisfy the stability constraint. This will be done computing sample posterior means as well as sample MAP. 
%\begin{itemize}
%\item \vspace{-.2cm} Draw samples $f^{(i)},g^{(i)}$, $i=1,..,N$ from the ``stable'' posterior distribution \eqref{eq: stable posterior general form}
%\item Using these samples compute  estimates $\hat P,\hat H$ and $\hat f,\hat g$ which satisfy the stability constraint. This will be done computing sample posterior means as well as sample MAP. 
%\end{itemize}
\end{problem}

In order to sample from the stable posterior \eqref{eq: stable posterior general form} one can use a Metropolis-Hasting type of algorithm as in \cite{WG-SR-DS:96}.
%\begin{enumerate}
%\item Set $k=1$ and initialize $(f^{(1)},g^{(1)})$ so that $f^{(1)}$ corresponds to a stable $A(z)$.
%\item Generate proposal moves $(f^{(k)},g^{(k)}) \rightarrow  (f^{'},g^{'})$ sampling from a proposal distribution
%$
%Q((f^{'},g^{'})| (f^{(k)},g^{(k)}))
%$
%\item Accept the move with probability 
%\begin{equation}\label{accept_pS}
%\alpha:=\min \left(1,\frac{p_S(f^{'},g^{'}|y) Q((f^{(k)},g^{(k)})| (f^{'},g^{'}))}{p_S(f^{(k)},g^{(k)}|y) Q((f^{'},g^{'})| (f^{(k)},g^{(k)}))}\right)
%\end{equation}
%\item If \emph{accepted} set $(f^{(k+1)},g^{(k+1)}) = (f^{'},g^{'})$, otherwise set $(f^{(k+1)},g^{(k+1)}) = (f^{(k)},g^{(k)})$. 
%\item $k \rightarrow k+1$ and go back to $2)$.
%\end{enumerate}

We have now to address two fundamental issues for this algorithm to be implementable, namely:
\begin{itemize}
\item[(i)] Design the proposal density $Q_{f,g}(\cdot| \cdot)$
%$Q((f^{'},g^{'})| (f^{(i)},g^{(i)}))$, where $(f^{'},g^{'})$ are the new candidate sample.
\item[(ii)] Compute the posterior $p_S(f,g|y)$, up to a constant multiplicative factor\footnote{This is because only  ratios of probabilities need to be computed.}.% in \eqref{accept_pS}
\end{itemize}

A preliminary step for both items (i) and (ii) is the computation of a set of samples $\eta_i \sim p(\eta|y)$ from the posterior of the hyperparameters, without accounting for the stability constraint. 

In the next subsections we address these three issues.
%We shall now address these three issues in the next  subsections

\subsection*{Sampling from the posterior density $p(\eta |y)$}
First, our aim is to  draw points from the  posterior density of $\eta$ given $y$. Notice that:
\begin{equation}
\label{eq: marginal posterior hyper-parameters}
p(\eta\vert y) = \frac{p_\eta(y)p(\eta)}{p(y)}
\end{equation}
where, as mentioned earlier on, $p(\eta)$ is assumed to be a non informative prior distribution, and $p(y)$ is the normalization constant. The marginal density $p_\eta(y)$ of $y$ given $\eta$ can be computed in closed form, as discussed in   \cite{GP-AC-GdN:11} and is given by \begin{equation}
\label{eq: marginal log likelihood}
p_\eta(y)= {\rm exp}\left( -\frac{1}{2} \ln(\det[2\pi\Sigma_\eta])- \frac{1}{2}y^T\Sigma_\eta^{-1}y\right)
\end{equation}
where
\begin{equation}\label{Sigma_eta}
\Sigma_\eta = AK_\eta A^\top + BK_\eta B^\top + \sigma^2I 
\end{equation}
where $\sigma^2:=Var\{e(t)\}$ is the  variance of the innovation process \eqref{IOmodel} and $A,B$ are matrices built with the past input-output data, see \cite{GP-AC-GdN:11} for details.

In order to obtain samples from  \eqref{eq: marginal posterior hyper-parameters} we implemented a  Metropolis-Hasting algorithm, see e.g. \cite{WG-SR-DS:96}. We are using a symmetric proposal distribution $q_\eta(\cdot \vert \cdot)$ which describes a random walk in the hyperparameter space, whose mean is centered in the present value and its variance contains information about the local curvature of the target. To do so, let us define:
\begin{equation}
\label{eq: initial hyper-parameters empirical bayes}
\begin{array}{ccl}
\underline{\eta} &=& \arg \min_\eta -\ln[p_\eta(y)p(\eta)]\\
H &=& -\frac{d^2 \ln[p_{\underline\eta}(y)p(\underline{\eta})]}{d\eta d\eta^T}
\end{array}
\end{equation}
that is the Hessian matrix computed in $\underline{\eta}$. Thus we define $q_\eta(\cdot \vert \mu) = \mathcal{N}(\mu,\gamma H^{-1})$ where $\gamma$ is a positive scalar chosen to obtain an acceptance probability in the MCMC algorithm around the $30\%$ via a pilot analysis, see e.g. \cite{GR-AG-WG:97}.\\
%In cases where the covariance $\gamma H^{-1}$ results to be inefficient to explore the support it can be exchanged by computing the sample covariance from the samples accepted during an additional pilot analysis (e.g. starting from a diagonal covariance, obtain a certain amount of samples and compute the sample covariance of these samples).\\
The acceptance rate of the MCMC results to be:
\begin{equation*}
\alpha_{\eta_{i}} = \min \left(1,\frac{p_{\eta_{i}}(y)p(\eta_i)}{p_{\eta_{i-1}}(y)p(\eta_{i-1})}\right)
\end{equation*}
%So far, we have characterized the MCMC to obtain hyperparameters distributed as a sampled form\eqref{eq: marginal posterior %hyper-parameters}. Now, exploiting the hyperparameters samples we can show how to design the second MCMC to obtain a %sampled version of $p(f_{\eta},g_{\eta}\vert\eta,y)$ in order to have \eqref{eq: stable posterior general form} eventually.\\

\subsection*{Proposal density}
It is well known in the MCMC literature that an accurate choice of the proposal distribution may have a remarkable impact on the performance of the Markov Chain. In this paper we adopt a data-driven proposal computed from the posterior distribution disregarding the stability constraint. The algorithm we consider is based on the approximation 
\begin{equation}\label{ApproxPosterior}
\hspace*{-2.5mm} p(f,g|y) \hspace*{-1mm} = \hspace*{-1mm} \int_\eta p(f,g|y,\eta) p(\eta|y)\, d\eta \simeq \frac{1}{N} \sum_{i=1}^{N}p_{\eta_i}(f,g|y)
\end{equation}
where $\eta_i$, $i=1,..,N$ are the samples from $p(\eta|y)$ drawn by the  MCMC algorithm  above and 
\begin{equation}\label{cond_posterior}
p_{\eta_i}(f,g|y) \sim {\cal N}\left(\mu_{\eta_i}^{MAP},\Sigma_{\eta_i}^{MAP}\right)
\end{equation}
is the (Gaussian) posterior density of $f,g$  when the  hyperparameters are fixed equal to $\eta_i$. The posterior means and variance are, respectively:
$\mu_\eta^{MAP}:=(\E_\eta[f \vert  y] , \E_\eta[g \vert, y])$ 
\begin{equation}
\label{eq: MAP conditional hyperparamters}
\begin{array}{ccl}
\E_\eta[f \vert  y]\hspace*{-2mm} &=& \hspace*{-2mm}K_\eta A^\top \Sigma_\eta^{-1}y , \quad
 \E_\eta[g \vert y] = K_\eta  B^\top \Sigma_\eta^{-1} y  \\
 \Sigma_\eta^{MAP} \hspace*{-2mm} &=& \hspace*{-2mm}
\underline K_\eta - \underline K_\eta
\begin{bmatrix}
A^\top \\ B^\top
\end{bmatrix}
%\left[\begin{array}{c} A^\top \\ B^\top \end{array}\right]
\Sigma_\eta^{-1}
%\left[\begin{array}{cc} A & B \end{array}\right]
\begin{bmatrix}
 A & B
\end{bmatrix}
\underline K_\eta\\
\underline K_\eta \hspace*{-2mm} &=& \hspace*{-2mm} \begin{bmatrix}
K_\eta & \underline{O}\\
\underline{O} & K_\eta 
\end{bmatrix}
 \end{array}
\end{equation}
and $\Sigma_\eta$ is defined in \eqref{Sigma_eta}.

From \eqref{ApproxPosterior} it follows that, in order to sample from the proposal density $p(f,g|y)$ one can
\begin{enumerate}
\item Sample $\eta_i \sim p(\eta|y)$
\item Sample $(f,g) \sim p_{\eta_i}(f,g|y)$ in \eqref{cond_posterior}
\end{enumerate}

\subsection*{Evaluation of the stable posterior $p_S(f,g|y)$}

The stable posterior in equation \eqref{eq: stable posterior general form} can be approximated as follows:
\begin{equation}\label{approx:stable:posterior}
\begin{array}{rcl}
\hspace*{-3mm} p_S(f,g\vert y) \hspace*{-1mm} &=& \hspace*{-2mm} \int p_S(f,g,\eta \vert y)\, d\eta \\
%\hspace*{-1mm} \hspace*{-2mm} &=& \hspace*{-1mm} \hspace*{-1mm} \frac{1}{p(y)}\int p(y|f,g)p_S(f,g|\eta) p(\eta)\, d\eta  \\
\hspace*{-2mm} &=& \hspace*{-1mm} \frac{1}{p(y)}\int p(y|f,g)p_S(f,g|\eta) p(\eta)\frac{q(\eta)}{q(\eta)}\, d\eta  \\
\hspace*{-2mm} &\simeq& \hspace*{-1mm} \frac{1}{N p(y)} \sum_{i=1}^{N} \frac{p(y|f,g)p_S(f,g|\eta_i) p(\eta_i)}{q(\eta_i)} 
\end{array}
\end{equation}
with $\eta_i \sim q(\eta)$. Note that the quantities $p(y|f,g)$, $p_S(f,g|\eta)$ and $p(\eta)$ can be evaluated. Thus, setting $q(\eta):=p(\eta|y)$ and using the MCMC algorithm described above to obtain samples from the posterior $p(\eta|y)$, the stable posterior $p_S(f,g\vert y) $ can then be approximated  (up to the irrelevant normalization constant $p(y)$) from equation  \eqref{approx:stable:posterior}.

\subsection*{Algorithm}
We are now ready to provide the MCMC algorithm to sample from the stable posterior $p_S(f,g|y)$ \eqref{eq: stable posterior general form}:
%%%%%%%%%%%%%%%%%%%%%%%%%%%%%%%%%%%%%%%%%%%%%%%%%%%%%%%%%%%

%%%%%%%%%%%%%%%%%%%%%%%%%%%%%%%%%%%%%%%%%%%%%%%%%%%%%%%%%%%%%%%%%%%
\begin{alg}[MCMC]
\label{alg: MCMC}
\mbox{ }\\
Hyper-parameters MCMC: 
\begin{enumerate}
\item Initialization: set $\eta_0 =\underline{\eta}$ using \eqref{eq: initial hyper-parameters empirical bayes}
\item For $i>0$ Iterate:
\begin{itemize}
\item Sample $\eta$ from $q_\eta(\cdot \vert \eta_{i-1}) \thicksim \mathcal{N}(\eta_{i-1},\gamma H^{-1}))$
\item Sample $u$ from a uniform distribution on $[0,1]$
\item Set 
$\eta_i = 
\left\lbrace
\begin{array}{ll}
\eta & \mbox{ if } u \leq \frac{p_{\eta_i}(y)p(\eta_i)}{p_{\eta_{i-1}}(y)p(\eta_{i-1})}\\
\eta_{i-1} & \mbox{ otherwise}
\end{array}
\right.$
\end{itemize}
\item After a burn-in period, keep the last $N$ samples $\eta_i$ which are (approximately) samples from   $p(\eta\vert y)$.
\end{enumerate}
%%%%%%%%%%%%%%%%%%%%%%%%%%%%%%%%%%%%%%%%%%%%%%%%%%%%
Predictor Impulse Responses MCMC:
\begin{enumerate}\addtocounter{enumi}{3} 
\item Initialization: compute $[f_0,g_0]$ from $\eta_0$ using~\eqref{cond_posterior}
\item For $i = 1$ to $N$ do
\begin{itemize}
%\drmargin{qui avrei imposto anche che se è la prima volta che vedo l'iperparametro il sample [f,g] è proprio E[f,g|y,eta], lo dico?}
\item  compute $\mu_{\eta_i}^{MAP}, \, \Sigma_{\eta_i}^{MAP}$  as in \eqref{eq: MAP conditional hyperparamters}
\item Sample $(f^{'},g^{'})$ from $\mathcal{N} (\mu_i^{MAP},\Sigma_i^{MAP})$
\item Compute  $\alpha$ as 
$$
\alpha:=\min \left(1,\frac{p_S(f^{'},g^{'}|y) p(f^{(k)},g^{(k)}|y)}{p_S(f^{(k)},g^{(k)}|y) p(f^{'},g^{'}|y)}\right)
$$
with $p_S(f,g|y)$ and $p(f,g|y)$ approximated as in \eqref{approx:stable:posterior} and in \eqref{ApproxPosterior}.
\item Sample $u$ from a uniform distribution on $[0,1]$
\item 
$\mbox{Set: } (f^{(i)},g^{(i)}) = 
\left\lbrace
\begin{array}{ll}
(f^{'},g^{'}) \hspace*{-4mm}& \mbox{ if } u \leq \alpha \\
(f^{({i-1})},g^{(i-1)})\hspace*{-4mm} & \mbox{ otherwise}
\end{array}
\right.$
\end{itemize}
%\item A sampled version of $p_S(f,g\vert y)$ is obtained
\item The samples $(f^{(i)},g^{(i)})$ obtained above are i.i.d. samples from $p_S(f,g|y)$ as requested by Problem~\ref{prb: MCMC}. 
The estimates of $P(z)$ and $H(z)$ can be obtained as:
\begin{itemize}
\item {\bf Minimum Variance Estimate}: from each sample $(f^{(i)},g^{(i)})$ compute the impulse responses $P_i(z)$ and $H_i(z)$ in \eqref{eq: system estimated from the predictor} and compute the averages  
\begin{equation}
\label{eq: MCMC posterior mean estimate}
\hspace*{-3mm}\hat{P}(z) = \frac{1}{N}\sum_{i=1}^NP_i(z), \,\, \hat{H}(z) = \frac{1}{N}\sum_{i=1}^NH_i(z)
\end{equation}
We shall define $\hat p := \{\hat p_k\}_{k \in [1,p]}$, $\hat h := \{\hat h_k\}_{k \in [1,p]}$ the inverse ${\cal Z}$-transforms of $\hat P$ and $\hat H$ in \eqref{eq: MCMC posterior mean estimate}.
\item {\bf Maximum a Posteriori Estimate}
\begin{equation}
\label{eq: MCMC map estimate}
\bar{f},\bar{g} = \arg \max_{f_i,g_i} p_S(f,g\vert y)
\end{equation}
%in the remaining of the paper the model obtained by \eqref{eq: system estimated from the predictor} using \eqref{eq: MCMC map estimate} will be called ``MCMC MAP'' model.
\end{itemize}
\end{enumerate}
\end{alg} 
In the remaining of the paper the model obtained by \eqref{eq: system estimated from the predictor} using \eqref{eq: MCMC posterior mean estimate} and \eqref{eq: MCMC map estimate} will be called ``MCMC posterior mean'' model ``MCMC MAP'' model, respectively.
Note that, from  \eqref{eq: MCMC posterior mean estimate}, an estimate of $P(z)$ is obtained directly. This is to guarantee that $P(z)$ is stable since the average $\sum_{i} \hat P_i(z)$ of BIBO stable function is BIBO stable. On the other hand, if one averaged\footnote{Recall that the average of stable polynomial is not necessarily a stable polynomial unless the degree is smaller than $3$, see \cite{MG:07}.} the $f^{(i)}$ directly, there would be no guarantee that the average $f$ would lead to a stable $A(z)$ (and thus a stable model). Of course, if needed,  an estimate of $F$ can be obtained from \eqref{predictor}  using $\hat P$ and $\hat H$ in \eqref{eq: MCMC posterior mean estimate}  :
$$
%\begin{array}{c}
\hat G(z): = \hat H^{-1}(z) \hat P(z), \qquad \hat F(z):= 1- \hat H^{-1}(z)
%\end{array}
$$

%Therefore for every $f_i$ we had to compute the impulse response $p_i$ and make an average as in \eqref{eq: MCMC posterior mean estimate} since it holds that %the average of stable rational functions is still stable.

%\begin{table}[hb]
%\begin{center}
%\caption{Margin settings}\label{tb:margins}
%\begin{tabular}{cccc}
%Page & Top & Bottom & Left/Right \\\hline
%First & 3.5 & 2.5 & 1.5 \\
%Rest & 2.5 & 2.5 & 1.5 \\ \hline
%\end{tabular}
%\end{center}
%\end{table}
% -------------------------------------------------------------------------------------- %
\section{Simulations}
\label{Section: Simulations}
% -------------------------------------------------------------------------------------- %
The performance of the 
%four\footnote{Recall that in  \ref{Section: MCMC } two estimators (MAP and Posterior Mean) are considered.} 
techniques described the paper 
%in Section \ref{Section: Stabilization with LMI}, %\ref{Section: Stabilization with Smooth Barrier} and \ref{Section: MCMC } 
are compared by means of a Monte Carlo experiment, considering identification or 
% At each run the aim is to estimate stable ARX models from an identification data set generated from a 
marginally stable models, i.e., with poles close to the complex unit circle.
%each time this model results unstable the techniques described in this paper are applied. 
%We shall consider  both in terms of quality of prediction of new data as well as on the reconstruction of   the ``true'' impulse response. 
%
%\subsection{Generation of Marginally Stable Model}
At each Monte Carlo run a $2^{nd}$-order SISO ARMAX model, called $M$, is generated:
\begin{equation}
\label{eq: Model}
A(z)y(t)= k z^{-1} B(z)u(t) + C(z)e(t)
\end{equation} 
The two complex conjugate roots of the monic polynomial $A(z)$ are placed in $0.996\cdot\exp(\pm j\frac{\pi}{3})$, $B(z)$ is a random polynomial whose roots are restricted to lie inside the circle of radius $0.9$ and $C(z)$ has randomly roots chosen in the interval $[0.65,0.73]$ so to ensure that the predictor impulse responses decay in no more then 30 steps.\\
The system input $u(t)$ and the disturbance noise $e(t)$ are independent white noise with unit variance (for both identification and test data sets). The constant $k$ is designed so that the signal-to-noise ratio of \eqref{eq: Model} is one.  More specifically,  let  $y_u(t):=B(z)/A(z) u(t)$ and $y_e(t):=C(z)/A(z)e(t)$, then $k$ as been set to: $k = \sqrt{var(y_e)/var(y_u)}$.
%
%\subsection{Experiment description}
A Monte Carlo  study of 5000 runs is implemented. At each run a model as \eqref{eq: Model} is used to generate an identification set of 400 samples and a test set of 1000 samples.\\
The predictor impulse responses $f$ and $g$ are estimated via the Bayesian System Identification described in \cite{GP-AC-GdN:11} which is based on the Stable Spline Kernel as a priori covariance and the hyperparameters %$\{\lambda_f,\lambda_g,\beta\}$ 
are determined as in \eqref{maxML}. The predictor impulse responses are negligible for time lags larger than $30$ and thus the truncation length is chosen  as $p= 30$. The variance of the noise $\sigma$ is computed via a low bias  Least Square identification method. 
The estimators $\hat P$ and $\hat H$ in \eqref{eq: system estimated from the predictor} obtained from the Stable Spline estimators $\hat f$, $\hat g$ ended up being unstable about $150$ times out of $5000$ Monte Carlo runs. In these cases the stabilization procedures described in these paper have been applied.
Thus our Monte Carlo analysis is limited to these $150$ data sets which resulted in unstable systems.

The \textit{CVX toolbox}, \cite{MG-SB-YY:06}, which is based on \textit{YALMIP}, was used in Matlab to solve the convex optimization problem \eqref{eq: LMI optimization problem first step}, with solver SeDuMi, \cite{JSS:01}. Instead, the Matlab function \textit{`fminsearch.m'} has been used to solve problem \eqref{eq: min lml + penalty function}.\\
Notice that all these unstable models have been stabilized by our techniques.
\subsection{Performance results}
In order to illustrate the identification performances, we first consider dominant poles of the estimated, which are shown 
\begin{figure}[!h]
\includegraphics[width=.9\columnwidth]{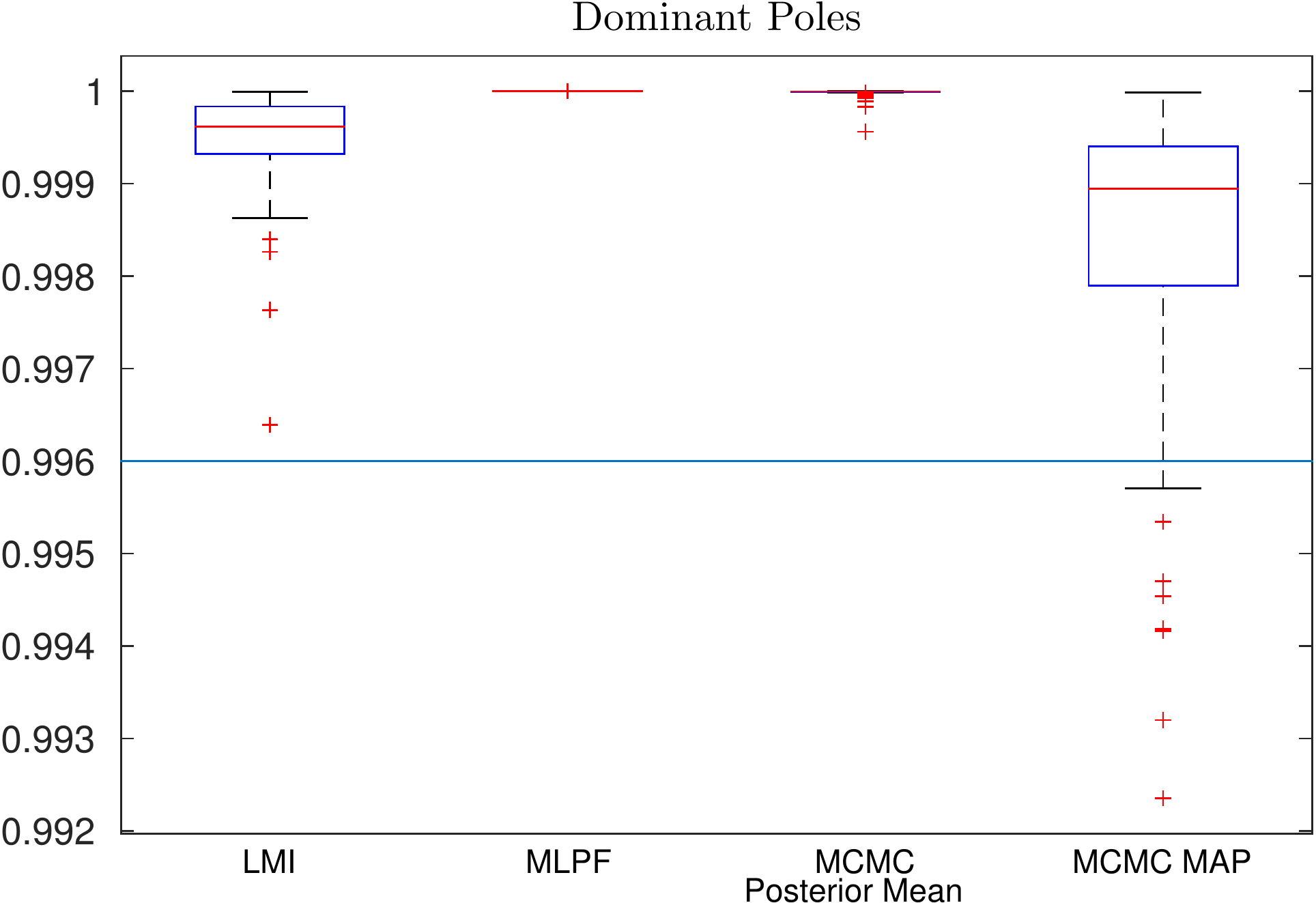}
\caption{Monte Carlo results. Boxplots of the absolute value of the dominant poles of the identified models. The horizontal line represents the absolute value of the dominant pole of the true model.}
\label{fig: Absolute value poles}
\end{figure}
in Figure~\ref{fig: Absolute value poles};   the horizontal line in 0.996 indicates the absolute value of the ``true''  dominant poles. All the estimation methods, and in particular ``ML+PF'' and ``MCMC posterior mean'',   tend to place the poles close to the  unit circle. 

%The situation is slightly different when comparing the performance 

In addition the estimated impulse responses are  compared to the ``true'' ones
in terms of relative errors on the estimated impulse responses:
\begin{equation}
\mbox{err}_i = \frac{1}{2}\frac{\Vert p_k - \hat{p}_k \Vert_2}{\Vert p_k\Vert_2}+\frac{1}{2}\frac{\Vert h_k - \hat{h}_k \Vert_2}{\Vert h_k\Vert_2} %-  \frac{\sum_1^{1000} h(t)}{1000} \Vert
\end{equation}
where $\{\hat{p}\}$ and $\{\hat{h}\}$ are the estimators of the true impulsed responses $\{p\}$ and $\{h\}$.
%

% one-step-ahead prediction $\hat{y}(t\vert t-1)$ is computed and it is compared to the new data given in the test set $y_{test}(t)$. At the $i$-th Monte Carlo run the one-step-ahead coefficient of determination is computed as:
%\begin{equation}
%COD_i = 100\left(1- \frac{\Vert y_{test}(t) - \hat{y}(t) \Vert_2}{\Vert y_{test}(t)\Vert_2} \right) %- \frac{\sum_1^{1000} y_{test}(t)}{1000} \Vert}
%\end{equation}

%It is interesting to observe that all models performs well in terms of prediction, all being comparable to the ``ideal'' predictor (i.e., that obtained using the ``true'' system).
% 
%\begin{figure}[!h]
%\begin{center}
%\includegraphics[width=\columnwidth]{Images/Compare_COD1-crop.pdf} 
%\caption{Monte Carlo results. Boxplots of the $\{COD_i \}$ in prediction on new data for the ``true'' and the identified models.}
%\label{fig: COD1 comparision}
%\end{center}
%\end{figure}
%
%Figure~\ref{fig: COD1 comparision} displays the boxplots of $\{COD_i \}$ for the $4$ possible identified models as well as for ``true''  model, $M$. \\ 
%The $\{COD_i\}$ for the ``true'' model are used as a reference in the prediction performance and the comparison shows that all the model identified perform remarkably well, with no %significant differences both in terms of median and variance. 
%The perturbation introduced by the stabilization techniques into the model identified with only the stable spline method seems not to affect negatively the property of %minimizing the prediction error \eqref{PE}.

\begin{center}
\begin{figure}[!h]
\includegraphics[width=.9\columnwidth]{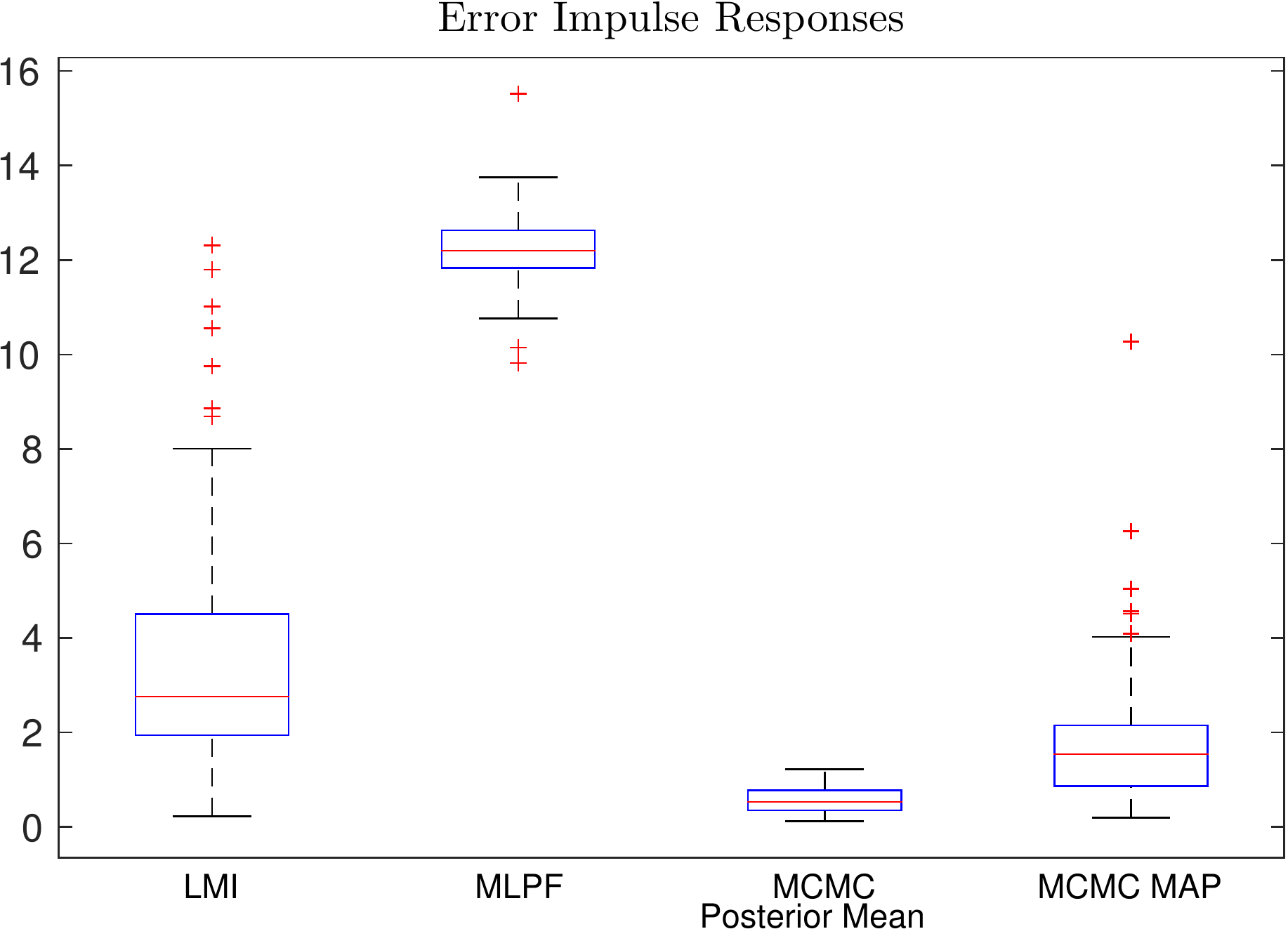}
\caption{Monte Carlo results. Boxplots of the $\{\mbox{err}_i\}$. }%Quantity to qualify the reconstruction of the system impulse responses.}
\label{fig: Error impulse response comparison}
\end{figure} 
\end{center}
Figure~\ref{fig: Error impulse response comparison} reports the Boxplots of $\{\mbox{err}_i\}$ for the estimated models. The  ``MCMC posterior mean'' estimator outperforms all the others significantly. The remaining three techniques yield rather poor quality properties in the identification of the system which is due to poor  estimation of ``dominant'' modes. Indeed, a higher absolute value of the dominant pole corresponds to a slower decay rate of the impulse responses. \\
When the estimators place a dominant pole very close to the unit circle, this results in a significant tail in the impulse response which, in turn yield a very high relative error. The algorithm ``MCMC posterior mean'' deserves a  separate discussion. In this case, since the estimated $P(z)$ is the average of all $\hat P_i's$, the dominant pole of $\hat P$ is the slowest among the dominant poles of $\hat P_i 's$. Yet, the effect of these dominant modes on the relative error is mitigated by 
 the factor $\frac{1}{N}$ in the average \eqref{eq: MCMC posterior mean estimate}.

\section{Conclusion}
\label{Section: Conclusions}
We have presented four different techniques to face the problem of identifying a stable system using a Bayesian framework based on the minimization of the predictor error. The experiment shows all methods ultimately produce stable models which perform comparably in terms of prediction error (not reported for reasons of space); however, 
%that the proposed techniques perform remarkably well in the prediction of new data. Whereas in the capability of reconstructing the system impulse responses. Only 
only the model estimated with the so called ``MCMC posterior mean'' technique perform satisfactorily in terms of impulse response fit.  
%The reasons being that the dominant poles of the model are often estimated  very close to the unit circle, thus inducing an excessively slow decay of  the impulse responses.\\
In future work, we will discuss new techniques to overcome the problem in the identification performance without the usage of a MCMC. In particular, we are looking for new  regularization which take in account penalty term both in the predictor and in the system impulse responses.

\section*{Acknoledgemnt}
This work has been  supported by MIUR through the  FIRB project ``Learning meets time'' (RBFR12M3AC)
%%%%%%%%%%%%%%%%%%%%%%%%%%%%%%%%%%%%%%%%%%%%%%%%%%%%%%%%%%%%%%%%%%%%%%%%%%%%%%%%

\bibliographystyle{IEEEtran}
\bibliography{Main}

\begin{thebibliography}{10}
\providecommand{\url}[1]{#1}
\csname url@rmstyle\endcsname
\providecommand{\newblock}{\relax}
\providecommand{\bibinfo}[2]{#2}
\providecommand\BIBentrySTDinterwordspacing{\spaceskip=0pt\relax}
\providecommand\BIBentryALTinterwordstretchfactor{4}
\providecommand\BIBentryALTinterwordspacing{\spaceskip=\fontdimen2\font plus
\BIBentryALTinterwordstretchfactor\fontdimen3\font minus
  \fontdimen4\font\relax}
\providecommand\BIBforeignlanguage[2]{{%
\expandafter\ifx\csname l@#1\endcsname\relax
\typeout{** WARNING: IEEEtran.bst: No hyphenation pattern has been}%
\typeout{** loaded for the language `#1'. Using the pattern for}%
\typeout{** the default language instead.}%
\else
\language=\csname l@#1\endcsname
\fi
#2}}

\bibitem{Ljung}
L.~Ljung, \emph{System Identification, Theory for the User}.\hskip 1em plus
  0.5em minus 0.4em\relax Prentice Hall, 1997.

\bibitem{Soderstrom}
T.~S{\"o}derstr{\"o}m and P.~Stoica, \emph{System Identification}.\hskip 1em
  plus 0.5em minus 0.4em\relax Prentice-Hall, 1989.

\bibitem{GP-AC-GdN:11}
G.~Pillonetto, A.~Chiuso, and G.~D. Nicolao, ``Prediction error identification
  of linear systems: a nonparametric gaussian regression approach,''
  \emph{Automatica}, no.~47, pp. 291--305, 2011.

\bibitem{SoedestroemLG1976}
T.~S\"{o}derstr\"{o}m, L.~Ljung, and I.~Gustafsson, ``Identifiability
  conditions for linear multivariable systems operating under feedback,''
  \emph{IEEE Trans. on Aut. Contr.}, vol.~21, pp. 837--840, 1976.

\bibitem{NgGA1977}
T.~Ng, G.~Goodwin, and B.~Andersson, ``Identifiability of mimo linear dynamic
  systems operating in closed loop,'' \emph{Automatica}, vol.~13, pp. 477--485,
  1977.

\bibitem{Tikhonov}
A.~Tikhonov and V.~Arsenin, \emph{Solutions of Ill-Posed Problems}.\hskip 1em
  plus 0.5em minus 0.4em\relax Washington, D.C.: Winston/Wiley, 1977.

\bibitem{SS2010}
G.~Pillonetto and G.~{De Nicolao}, ``A new kernel-based approach for linear
  system identification,'' \emph{Automatica}, vol.~46, no.~1, pp. 81--93, 2010.

\bibitem{Aronszajn50}
N.~Aronszajn, ``Theory of reproducing kernels,'' \emph{Trans. of the American
  Mathematical Society}, vol.~68, pp. 337--404, 1950.

\bibitem{Rasmussen}
C.~Rasmussen and C.~Williams, \emph{{G}aussian Processes for Machine
  Learning}.\hskip 1em plus 0.5em minus 0.4em\relax The MIT Press, 2006.

\bibitem{CM-GP:96}
M.~Chilali and P.~Gahinet, ``H-infinity design with pole placement
  constraints:an lmi approach.'' \emph{IEEE Transactions on Automatic Control},
  vol.~41, pp. 358--367, 1996.

\bibitem{DNM-RAC:13}
D.~N. Miller and R.~A. de~Callafon, ``Subspace identification with eigenvalue
  constraints,'' \emph{Automatica}, vol.~49, pp. 2468--2473, 2013.

\bibitem{WG-SR-DS:96}
S.~R. W.Gilks and D.~Spiegehalter, \emph{Markov Chain Monte Carlo in
  Practice}.\hskip 1em plus 0.5em minus 0.4em\relax London: Chapman and Hall,
  1996.

\bibitem{GR-AG-WG:97}
G.~Roberts, A.~Gelkman, and W.Gilks, \emph{Weak convergence and optimal scaling
  of random walk metropolis algorithms}.\hskip 1em plus 0.5em minus 0.4em\relax
  Ann. Appl. Prob, 1997, vol.~7.

\bibitem{MG:07}
M.~Gora, ``Stability of the convex combination of polynomials,'' \emph{Control
  and Cybernetics}, 2007.

\bibitem{MG-SB-YY:06}
M.~Grant, S.~Boyd, and Y.~Ye, ``Disciplined convex programming,'' in
  \emph{Global Optimization: from Theory to Implementation, Nonconvex
  Optimization and Its Applications}, L.~Liberti and N.~Maculan, Eds.\hskip 1em
  plus 0.5em minus 0.4em\relax New York: Springer, 2006, pp. 155--210.

\bibitem{JSS:01}
J.~Sturm, ``Using sedumi, a matlab toolbox for optimization over symmetric
  cones,'' 2001.

\end{thebibliography}

%%%%%%%%%%%%%%%%%%%%%%%%%%%%%%%%%%%%%%%%%%%%%%%%%%%%%%%%%%%%%%%%%%%%%%%%%%%%%%%%

%\begin{appendix}
%
%\end{appendix}

\end{document}